\documentclass[sigconf]{acmart}

\usepackage{algorithmic}
\usepackage{algorithm}
\usepackage{amsmath}
\usepackage{booktabs}
\usepackage{tabularx}

\usepackage{amssymb}
\usepackage{enumitem}
\usepackage{bm}
\usepackage{etoolbox}
\usepackage{balance}
\usepackage{multirow}
\AtBeginDocument{%
  \providecommand\BibTeX{{%
    \normalfont B\kern-0.5em{\scshape i\kern-0.25em b}\kern-0.8em\TeX}}}

\copyrightyear{2024}
\acmYear{2024}
\setcopyright{acmlicensed}\acmConference[MM '24]{Proceedings of the 32nd ACM International Conference on Multimedia}{October 28-November 1, 2024}{Melbourne, VIC, Australia}
\acmBooktitle{Proceedings of the 32nd ACM International Conference on Multimedia (MM '24), October 28-November 1, 2024, Melbourne, VIC, Australia}
\acmDOI{10.1145/3664647.3681425}
\acmISBN{979-8-4007-0686-8/24/10}
\begin{document}

\title{Agent Aggregator with Mask Denoise Mechanism for
Histopathology Whole Slide Image Analysis}

\author{Xitong Ling}
\authornote{Both authors contributed equally to this research.}
\affiliation{%
  \institution{Tsinghua University}
  \city{Shenzhen}
  \country{China}
}
\email{lingxt23@mails.tsinghua.edu.cn}

\author{Minxi Ouyang}
\authornotemark[1]
\affiliation{%
  \institution{Tsinghua University}
  \city{Shenzhen}
  \country{China}
}
\email{oymx23@mails.tsinghua.edu.cn}

\author{Yizhi Wang}
\affiliation{%
  \institution{Tsinghua University}
  \city{Shenzhen}
  \country{China}
}
\email{yz-wang22@mails.tsinghua.edu.cn}

\author{Xinrui Chen}
\affiliation{%
  \institution{Tsinghua University}
  \city{Shenzhen}
  \country{China}
}
\email{cxr22@mails.tsinghua.edu.cn}

\author{Renao Yan}
\affiliation{%
  \institution{Tsinghua University}
  \city{Shenzhen}
  \country{China}
}
\email{yra21@mails.tsinghua.edu.cn}

\author{Hongbo Chu}
\affiliation{%
  \institution{Tsinghua University}
  \city{Shenzhen}
  \country{China}
}
\email{zhu-hb23@mails.tsinghua.edu.cn}

\author{Junru Cheng}
\affiliation{%
  \institution{Research Institute of Tsinghua}
  \city{Shenzhen}
  \country{China}
}
\email{cheng_jr@foxmail.com}

\author{Tian Guan}
\affiliation{%
  \institution{Tsinghua University}
  \city{Shenzhen}
  \country{China}
}
\email{guantian@sz.tsinghua.edu.cn}

\author{Sufang Tian}
\affiliation{%
  \institution{Wuhan University}
  \city{Wuhan}
  \country{China}
}
\email{sftian@whu.edu.cn}

\author{Xiaoping Liu}
\affiliation{%
  \institution{Wuhan University}
  \city{Wuhan}
  \country{China}
}
\authornote{Corresponding authors.}
\email{liuxiaoping@whu.edu.cn}

\author{Yonghong He}
\authornotemark[2]
\affiliation{%
  \institution{Tsinghua University}
  \city{Shenzhen}
  \country{China}
}
\email{heyh@sz.tsinghua.edu.cn}



\renewcommand{\shortauthors}{Xitong Ling et al.}

\begin{abstract}
Histopathology analysis is the gold standard for medical diagnosis. Accurate classification of whole slide images (WSIs) and region-of-interests (ROIs) localization can assist pathologists in diagnosis. The gigapixel resolution of WSI and the absence of fine-grained annotations make direct classification and analysis challenging. In weakly supervised learning, multiple instance learning (MIL) presents a promising approach for WSI classification. The prevailing strategy is to use attention mechanisms to measure instance importance for classification. However, attention mechanisms fail to capture inter-instance information, and self-attention causes quadratic computational complexity. To address these challenges, we propose AMD-MIL, an agent aggregator with a mask denoise mechanism. The agent token acts as an intermediate variable between the query and key for computing instance importance. Mask and denoising matrices, mapped from agents-aggregated value, dynamically mask low-contribution representations and eliminate noise. AMD-MIL achieves better attention allocation by adjusting feature representations, capturing micro-metastases in cancer, and improving interpretability. Extensive experiments on CAMELYON-16, CAMELYON-17, TCGA-KIDNEY, and TCGA-LUNG show AMD-MIL's superiority over state-of-the-art methods.

\end{abstract}

\ccsdesc[500]{Computing methodologies~Computer vision tasks}

\keywords{histopathology diagnosis, multiple instance learning, agent attention, mask denoise mechanism}



\maketitle
\begin{figure}[htpb]
\centering
\includegraphics[width=1\linewidth]{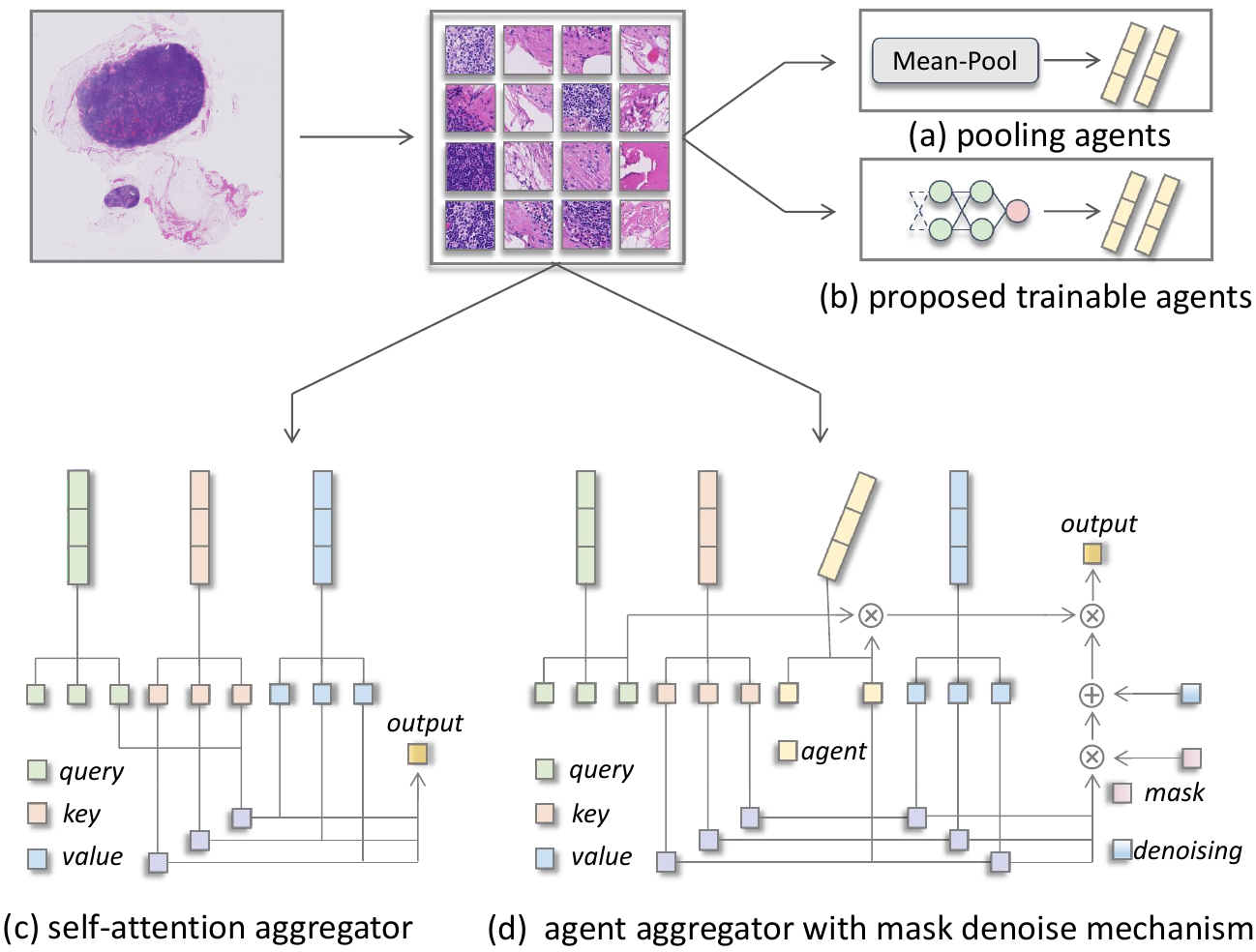}
\vspace{-4mm}
\caption{Comparison of core modules: (a) pooling agents. (b) proposed trainable agents. (c) self-attention mechanism. (d) proposed agent aggregator with mask denoise mechanism. Mask and denoising are learnable matrices.}
\label{F1}
\end{figure}

\section{Introduction}
The advancement of deep learning technologies and increased computational capacity have significantly enhanced the field of computational pathology~\cite{bera2019artificial,kather2019deep,song2023artificial,lu2021ai}. 
This progress assists physicians in diagnosis and standardizes pathological diagnostics~\cite{fu2020pan,moen2019deep}. However, analyzing histopathology whole slide images (WSIs) significantly differs from typical computer vision tasks~\cite{mckinney2020international}. 
A single WSI, with its gigapixel resolution, makes obtaining pixel-level annotations impracticable, in contrast to natural images~\cite{esteva2019guide}. 
The Multiple Instance Learning (MIL) method is currently the mainstream framework for analyzing histopathology slides using only WSI-level annotations~\cite{srinidhi2021deep,maron1997framework,xiang2022exploring}. 
MIL methods consider the entire WSI as a bag, with each patch within it as an instance~\cite{chen2022scaling,wang2018revisiting}. 
If any instance within the WSI is classified as cancerous, then the entire WSI is labeled as such~\cite{campanella2019clinical,wang2022weakly}. 
The WSI is labeled as normal only if all instances within it are normal.

Current MIL methods have two stages: segmenting the WSI into patches and using a pre-trained feature extractor to embed features. These features are then aggregated using methods such as mean-pooling, max-pooling, ABMIL~\cite{ilse2018attention}, DSMIL~\cite{li2021dual}, and TransMIL~\cite{shao2021transmil}, and then mapped for classification.

ABMIL~\cite{ilse2018attention} and DSMIL~\cite{li2021dual} use lightweight attention mechanisms for information aggregation. However, they overlook relationships between instances, hindering global modeling and capturing long-distance dependencies. TransMIL introduces self-attention~\cite{vaswani2017attention} within MIL's aggregator. Self-attention calculates relations between any two patches in a WSI, capturing long-distance dependencies. It also dynamically allocates weights based on input importance, enhancing the model's ability to process complex data. However, the quadratic complexity of self-attention challenges its application in MIL aggregators. TransMIL uses Nyström~\cite{xiong2021nystromformer} Self-attention instead. Nyström Self-attention selects a subset of elements, known as landmarks, to approximate attention scores. Nyström Self-attention down-samples query and key vectors locally along the instance token dimension. This approach has two main issues: sampling based on adjacent instances can dilute significant instance contributions, and the variance in instance numbers across bags requires padding during down-sampling. This can lead to aggregation imbalances and unstable outcomes.

To address the quadratic complexity issue of self-attention, TransMIL~\cite{shao2021transmil} employs Nystr\"om attention~\cite{xiong2021nystromformer} as the substitute for the standard self-attention module. Nystr\"om attention selects a subset of sequence elements, also known as landmarks, to approximate the attention scores for the entire sequence. Specifically, in the Nystr\"om attention mechanism, the local downsampling of query and key matrices is implemented along the dimension of the instance tokens. This approach has two significant issues. Firstly, since the sampling process relies on adjacent instances, many insignificant ones might dilute the impact of significant instances. Secondly, equidistant division is not always the optimal sampling strategy, as the distribution of information in a sequence may be uneven. Fixed sampling intervals might fail to capture all crucial information points, leading to a decrease in approximation quality.

To address these challenges, We transform the pooling agent into trainable matrices for effective mapping.
Furthermore, to indirectly achieve a more rational distribution of attention scores through adjustments in instance representations, we introduce the mask denoise mechanism for dynamic adaptation.

Agent attention~\cite{han2023agent} introduces the agent tokens in addition to query, key, and value tokens. Agent tokens act as the agent for the query tokens, aggregating information from the key and value tokens, and then information is returned to the query tokens via a broadcasting mechanism. Given the lesser number of agent tokens compared to sequence tokens, the agent mechanism can reduce the computational load of standard self-attention. However, agent tokens are obtained through mean pooling of the query tokens in standard agent attention, making it challenging to adapt to the variable-length token inputs of pathological multiple instance tasks. Additionally, mean pooling, by aggregating features through local averaging, may result in missing important information. Consequently, we adjust the number of agent tokens as a hyperparameter and substitute the mean pooling agent tokens with trainable agent tokens.

Moreover, we introduce the mask denoise mechanism to dynamically refine attention scores by adjusting instance representations. Mask and denoising matrices, matching the agent's aggregated value dimension, are generated by projecting this value through a linear layer. Mask matrices transform into binary matrices via threshold filtering, not directly from the value token but their high-level mapping, allowing dynamic adaptation to the input. Then, the mask directly multiplies with the value, filtering out non-significant representations. However, as the mask applies binary filtering to the value, it might suppress unimportant instances excessively, thereby introducing relative noise. Therefore, we introduce the denoising matrices from the agent-aggregated values to correct the relative noise.
We conducted extensive comparative experiments and ablation studies on four datasets to verify the effectiveness of the trainable agent aggregator and mask denoise mechanism.

\section{Related Work}
\begin{figure*}[htbp]
\centering
\includegraphics[width=1\linewidth]{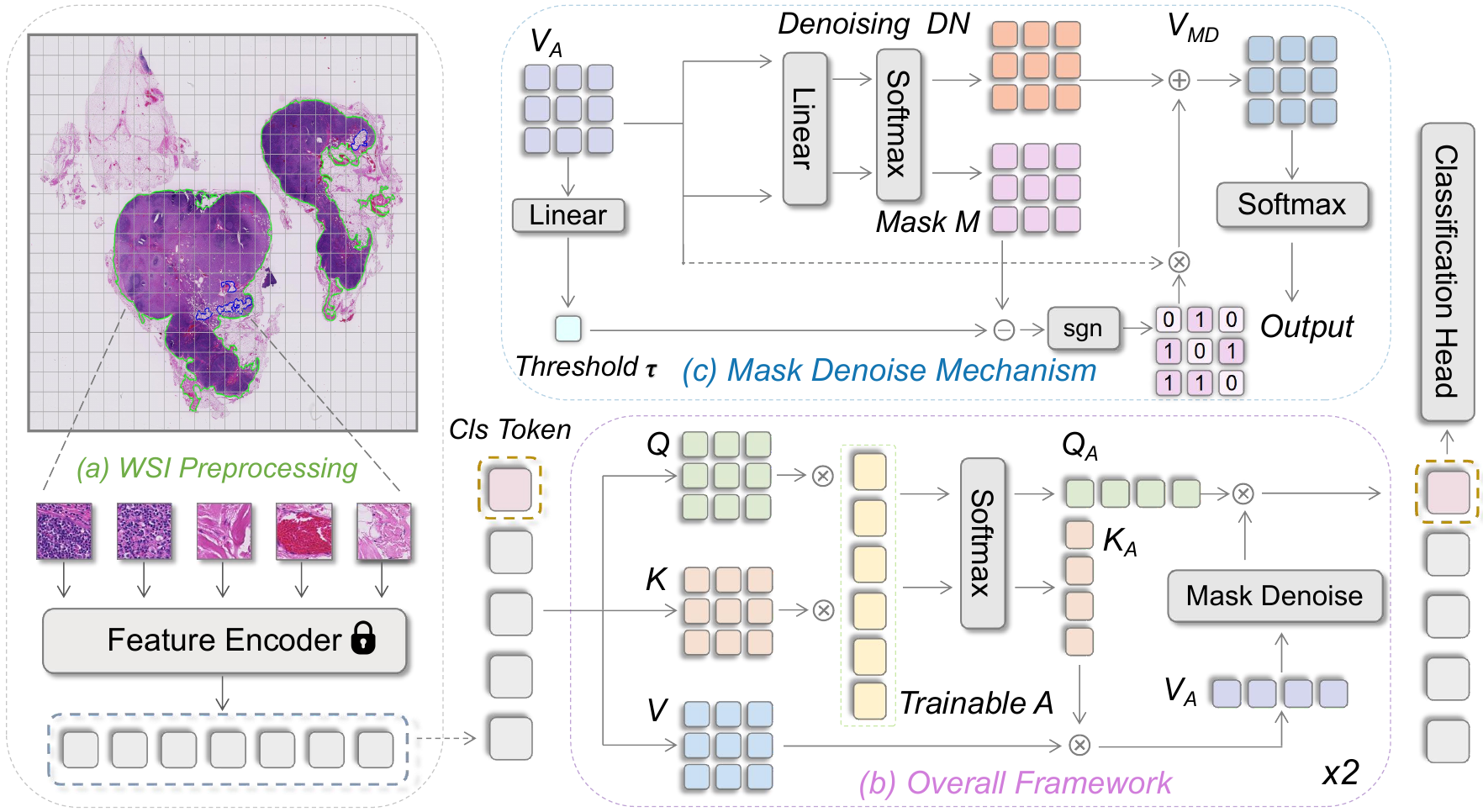}
\caption{Overall process: (a) the preprocess of WSI. (b) overall framework of AMD-MIL. (c) proposed mask denoise mechanism.}
\end{figure*}
\label{F2}
\subsection{Multiple Instance Learning for WSI Analysis}
MIL methods demonstrate significant potential in classifying and analyzing histopathology images. In this framework, a WSI is treated as a bag, and its local regions are instances. MIL paradigms are categorized into three types: instance-based, embedding-based, and bag-based methods. The instance-based method scores each instance and then aggregates these scores to predict the bag's label.

The embedding-based method employs a pre-trained feature encoder to generate instance representations, which are then aggregated for classification.
This enhances fit with Deep Neural Networks (DNN) but reduces the interpretability. Bag-based approaches classify by comparing distances between bags, with the main challenge being identifying a universal distance metric. 
Current MIL advancements focus on developing specialized feature encoders, enhancing aggregators, data augmentations, and improving training strategies.

Feature encoders pre-trained on natural images often struggle to extract high-level histopathology features, such as specific textures and morphological structures.
Transpath~\cite{wang2022transformer} trains A hybrid architecture of CNN and a swin-transformer feature encoder on one hundred thousand WSIs using a semantically relevant contrastive learning approach. 
IBMIL~\cite{Lin_2023_CVPR} also utilizes a feature encoder pre-trained on nine pathological datasets through a self-supervised method with MOCOv3~\cite{xinlei2021empirical}. 
Representations generated by these pathology-specific feature extractors significantly outperform those obtained from feature encoders pre-trained on ImageNet~\cite{deng2009imagenet} in downstream tasks. 

The most common aggregation strategies for instance-based and embedding-based methods include pooling and attention mechanisms.
Mean-MIL and Max-MIL aggregate representations and categorize through the average and maximum values respectively, but fixed aggregation mechanisms cannot adapt to varying inputs. 
In contrast, ABMIL employs attention mechanisms to aggregate features through trainable weights. 
Similarly, CLAM uses gated attention and a top-k selection strategy for bag label prediction. 
TransMIL, on the other hand, applies a linear approximation of self-attention to explore relationships between instances. 
WiKG~\cite{li2024dynamic} introduces a knowledge-aware attention mechanism, enhancing the capture of relative positional information among instances.
HAT-Net+~\cite{9858028} advances cell graph classification by leveraging a unique, parameter-free strategy to dynamically merge multiple hierarchical representations, effectively capturing the complex relationships and dependencies within cell graphs.

To enhance performance and stability, various methods employ data augmentation. For example, DTFD~\cite{zhang2022dtfd} increases the number of bags using a partitioning pseudo-bag split strategy and uses the double-tier MIL framework to use the intrinsic features.

In terms of training strategies, IBMIL~\cite{Lin_2023_CVPR} uses interventional training to reduce contextual priors. SSC-MIL~\cite{ye2023semantic} leverages semantic-similarity knowledge distillation to exploit latent bag information. MHIM-MIL~\cite{tang2023multiple} addresses key instances with hard example mining. LNPL-MIL~\cite{shao2023lnpl} proposes the training strategy of learning from noise pseudo labels, which can address the problem of semantical unalignment between instances and the bag. 

\subsection{Approximate Self-Attention Mechanism}
The self-attention~\cite{vaswani2017attention} mechanism is capable of grasping dependencies over long distances to facilitate comprehensive modeling, but its quadratic complexity limits the increase in input sequence length. Consequently, research on approximate self-attention mechanisms aims to reduce the complexity to linear while maintaining global modeling capability.

Nyström attention~\cite{xiong2021nystromformer} uses the Nyström method to estimate eigenvalues and eigenvectors, approximating self-attention~\cite{vaswani2017attention} by selecting a few landmarks, reducing computational and storage needs. Focused Linear attention~\cite{han2023flatten} uses nonlinear reweighting to focus on important features. Agent attention~\cite{han2023agent} introduces agents representing key information, significantly lowering computational complexity by computing attention only among these agents.

These advancements in approximate attention mechanisms provide a new perspective for enhancing aggregators in MIL methods.

\section{Methodology}

\subsection{MIL and Feature extraction}

In the MIL methodology, each WSI is conceptualized as a labeled bag, wherein its constituent patches are considered as instances possessing indeterminate labels. Taking binary classification of WSIs as an example, the input WSI $X$ is divided into numerous patches $\{(x_1, y_1),\cdots,(x_N,y_N)\}$, encompassing $N$ instances of $x_i$. 
Under the MIL paradigm, the correlation between the bag's label, $Y$, and the labels of instances ${y_i}$ is established as follows: 

\begin{align}
Y = 
\begin{cases} 
1, & \text{iff } \sum y_{i} > 0 \\
0, & \text{else}
\end{cases},
\end{align}

Given the undisclosed nature of the labels for the instances $y_{i}$, the objective is to develop a classifier, $\mathcal{M}(X)$, tasked with estimating $\hat{Y}$. 
In alignment with methodologies prevalent in contemporary research, the classifier can be delineated into three steps: feature extraction, feature aggregation, and bag classification. These processes can be defined as follows:

\begin{align}
\hat{Y} \leftarrow \mathcal{M}(X) := h(g(f(X))),
\end{align}
where $f$, $g$, and $h$ represent the feature extractor, feature aggregator, and the MIL classifier.  

The feature aggregator is considered to be the most important part of summarizing features, which can aggregate features of different patches. The attention mechanisms can discern the importance of patches in a WSI, and it is widely used in the feature aggregator. Attention-based and self-attention-based MIL are the main methods currently used.

In the attention-based MIL~\cite{zhang2022dtfd}, the  feature aggregator can be defined as:
\begin{align}
G = \sum_{i=1}^{N} a_i h_i = \sum_{i=1}^{N} a_i f(x_i) \in \mathbb{R}^D,\label{eq:state_space_model}
\end{align}
where $G$ is the bag representation, $h_i \in \mathbb{R}^D$ is the extracted feature for the patch $x_i$ through the feature extractor $f$, $a_i$ is the trainable scalar weight for $h_i$ and $D$ is the dimension of vector $G$ and $h_i$. 

In the self-attention-based~\cite{vaswani2017attention} MIL, the feature aggregator can be defined as:

\begin{align} 
Q = HW_Q,  K = HW_K,  V = HW_V, \label{eq:QKV}
\end{align} 
\begin{align} 
O = \text{softmax}\left(\frac{QK^T}{\sqrt{d_q}}\right) V = SV
, \label{eq:self-attention}
\end{align} 

where $W_Q$, $W_K$, and $W_V$ represent trainable matrices, $H$ denotes the collection of patch features, $O$ has integrated the attributes of the other features, and $d_q$ is the dimension of the query vector.

\subsection{Attention Aggregator}
During the computation of $\text{Sim}(Q, K)$ as defined in Eq. ~\ref{eq:self-attention}, the algorithmic complexity scales quadratically with $\mathcal{O}(N^2)$. Given that $N$ frequently comprises several thousand elements, this substantially extends the expected computational time. Linear attention offers a reduction in computational time but at the expense of information. To mitigate this issue, transmil~\cite{shao2021transmil} employs the Nystr\"om approximation for Eq. ~\ref{eq:self-attention} ~\cite{xiong2021nystromformer}. The matrices \( \tilde{Q} \) and \( \tilde{K} \) are constructed, and the mean of each segment is computed as follows:

\begin{align} 
\tilde{Q} &= [\tilde{q}_1; \ldots; \tilde{q}_m], & \tilde{q}_j &= \frac{1}{m} \sum_{i=(j-1)\times l+1}^{(j-1)\times l+m} q_i, & \forall j &= 1, \ldots, m \\
\tilde{K} &= [\tilde{k}_1; \ldots; \tilde{k}_m], & \tilde{k}_j &= \frac{1}{m} \sum_{i=(j-1)\times l+1}^{(j-1)\times l+m} k_i, & \forall j &= 1, \ldots, m
\end{align}
where \( \tilde{Q} \in \mathbb{R}^{m \times D} \) and \( \tilde{K} \in \mathbb{R}^{m \times D} \).

The approximation of the \( \hat{S} \) in Eq. ~\ref{eq:self-attention} can then be expressed as:
\begin{align}
\hat{S} = \text{softmax}\left(\frac{Q\tilde{K}^T}{\sqrt{d_q}}\right) Z^* \text{softmax}\left(\frac{\tilde{Q}K^T}{\sqrt{d_q}}\right),
\end{align}
where, \( Z^* \) represents the approximate solution to $z(\tilde{Q},\tilde{K},Z)=0$, necessitating a linear number of iterations for convergence.

In MIL tasks, Nystr\"om attention filters out patches with important features because of the sampling mechanism. Moreover, the difference in N will lead to an overall imbalance during local downsampling. So we consider agent attention methods with linear time complexity and the agent attention mechanism~\cite{han2023agent} can be written as: 
\begin{align} 
O = \sigma(QA^T) \sigma(AK^T) V, \label{eq:agent}
\end{align}
where $\sigma(\cdot)$ is the Softmax function, $Q,K,V$ are defined in equation Eq. ~\ref{eq:QKV}.
Here $A\in \mathbb{R}^{n \times D}$ is the agent matrix pooling from $Q$.
The term $D$ stands for the feature dimension, while $n$ refers to the agent dimension and acts as a hyperparameter.

Given that the agent is non-trainable and the distribution of attention scores may not be optimal, it becomes imperative to establish an adaptive agent capable of dynamically adjusting the attention score distribution to enhance model performance and flexibility.

\begin{algorithm}[!tbp]
    \caption{Agent Aggregator With Mask Denoise Mechanism}
    \label{alg:AOA}
    \renewcommand{\algorithmicrequire}{\textbf{Input:}}
    \renewcommand{\algorithmicensure}{\textbf{Output:}}
    \begin{algorithmic}[1]
        \REQUIRE H : ( B , N , D )     
        \ENSURE Y : ( B , N , D )      
        \STATE  $\textit{\textcolor[RGB]{110,154,155}{// \;H : bag\,\,features}}$
        \STATE  $\textit{\textcolor[RGB]{110,154,155}{// \;B : batch \;\; N : token\,\,number \;\; D : feature\,\,dimensions}}$
        \STATE  $Q , K , V$  : ( B , N , D )  $\longleftarrow$  nn.linear ( H )
        \STATE  $A$ : ( B , n , D )  $\longleftarrow$  trainable parameters
        \STATE  $\textit{\textcolor[RGB]{110,154,155}{// \;n : number\,\,of\,\,agent\,\,tokens}}$
        \STATE  $ Q_A $: ( B , N , n )   $\longleftarrow$  torch.matmul ( $Q$ , $A^T$ )
        \STATE  $ K_A $: ( B , n , N )   $\longleftarrow$  torch.matmul ( $A$ , $K^T$ )
        \STATE  $ V_A $ : ( B , n , D )  $\longleftarrow$  torch.matmul ( $K_A$ , $V$ )
        \STATE  $ M $ : ( B , n , D)  $\longleftarrow$  nn.linear ( $V_A$ )
        \STATE  $ THR $ : ( B , 1) $\longleftarrow$  nn.linear ( $V_A$ ) . suqeeze ( ) . mean ( -1 )
        \STATE  $ M_{t}$ : ( B , n , D)  $\longleftarrow$\text{torch.where} ( $M$ > $THR$ , 1 , 0 )
        \STATE $V_M$ : ( B , n , D ) $\longleftarrow$  torch.mul ( $V_A$ , $M_t$ )
        \STATE  $ DN $ : ( B , n , D)  $\longleftarrow$  nn.linaer ( $V_A$ )
        \STATE $V_{MD}$ : ( B , n , D ) $\longleftarrow$  torch.add ( ${V}_{M}$ , $DN$ )
        \STATE $Y$ : ( B , N , D )  $\longleftarrow$ torch.matmul( $Q_A$ , ${V}_{MD}$ )
        \STATE  $\textit{\textcolor[RGB]{110,154,155}{// \;Y : weighted\,\,bag\,\,features}}$
        \RETURN $Y$
    \end{algorithmic}
\end{algorithm}

\subsection{Agent Mask Denoise Mechanism}

As illustrated in Figure ~\ref{F1}, our overall framework is based on Eq. ~\ref{eq:self-attention} and Eq. ~\ref{eq:agent}. The proposed overall framework is shown in Figure ~\ref{F2}. Before the input features are processed by the model, a class token is embedded into them, resulting in the feature matrix $H \in \mathbb{R}^{D \times (N+1)}$, where $D$ is the dimension of the features and $(N+1)$ represents the number of patches, including the embedded class token.
\\ %
\textbf{Trainable Agent.} 
In the previously outlined methodology, matrix $A$ in Eq. ~\ref{eq:agent} is initially from matrix $Q$ through mean pooling, $A=pooling(Q) \in \mathbb{R}^{n \times D}$, indicating a limitation in encapsulating the entirety of information present within $Q$. To overcome this limitation, $A$ is defined as a trainable matrix.
Through matrix $A \in \mathbb{R}^{n \times D} $, the intermediate matrices $Q_A=QA^T \in \mathbb{R}^{(N+1) \times n}$ and $K_A=AK^T \in \mathbb{R}^{n \times (N+1)}$ can be obtained. Utilizing  the general attention strategy, the intermediate variable is:

\begin{equation}
\begin{aligned}
V_A &=  \sigma(K_A) V \\
&=  \sigma(AK^T) V  \in \mathbb{R}^{n \times D}, \label{eq:KV}
\end{aligned}
\end{equation}
\\ %

\begin{table*}[t]
\caption{
Performance of AMD-MIL on  CAMELYON-16, CAMELYON-17, TCGA-LUNG, and TCGA-KIDNEY datasets.}
\label{big table}
\centering
\renewcommand\arraystretch{1.2}
\resizebox{0.95\textwidth}{!}{
\begin{tabular}{ccccccccccccc}
\toprule
\multirow{2}{*}[-3pt]{Method} & \multicolumn{3}{c}{CAMELYON-16} & \multicolumn{3}{c}{CAMELYON-17} & \multicolumn{3}{c}{TCGA-LUNG} & \multicolumn{3}{c}{TCGA-KIDNEY} \\ \cmidrule(r){2-4} \cmidrule(r){5-7} \cmidrule(r){8-10} \cmidrule(r){11-13}
& ACC(\%) & AUC(\%) & F1(\%) & ACC(\%) & AUC(\%) & F1(\%) & ACC(\%) & AUC(\%) & F1(\%) & ACC(\%) & AUC(\%) & F1(\%) \\
\midrule
MeanMIL  
& 79.4\textsubscript{2.12} & 83.3\textsubscript{2.31} & 78.5\textsubscript{2.23} 
& 69.5\textsubscript{2.14} & 69.2\textsubscript{1.51} & 65.4\textsubscript{2.21} 
& 82.4\textsubscript{1.31} & 86.4\textsubscript{1.62} & 82.0\textsubscript{2.11} 
& 90.3\textsubscript{1.49} & 93.1\textsubscript{1.04} & 87.9\textsubscript{1.10}  \\
MaxMIL 
& 76.4\textsubscript{0.91} & 80.4\textsubscript{2.04} & 75.4\textsubscript{1.55} 
& 66.7\textsubscript{1.45} & 70.2\textsubscript{1.52} & 65.8\textsubscript{0.92} 
& 87.7\textsubscript{1.12} & 87.4\textsubscript{1.34} & 88.7\textsubscript{1.72} 
& 91.2\textsubscript{1.58} & 93.5\textsubscript{1.13} & 86.8\textsubscript{1.32}  \\
ABMIL \cite{ilse2018attention} 
& 84.8\textsubscript{1.14} & 85.9\textsubscript{1.05} & 84.1\textsubscript{1.22} 
& 78.7 \textsubscript{1.98} & 77.3 \textsubscript{1.64} & 75.3 \textsubscript{1.32} 
& 88.4\textsubscript{2.04} & 93.1\textsubscript{2.23} & 87.6\textsubscript{2.10}
& 91.6\textsubscript{0.93} & 94.1\textsubscript{0.82} & 88.5\textsubscript{1.23} \\
G-ABMIL  \cite{ilse2018attention} 
& 84.0\textsubscript{1.26} & 85.3\textsubscript{1.11} & 83.6\textsubscript{1.34} 
& 79.9\textsubscript{1.76} & 79.3\textsubscript{1.87} & 76.2\textsubscript{1.82} 
& 87.6\textsubscript{1.77} & 91.0\textsubscript{1.63} & 86.3\textsubscript{1.82} 
& 91.4\textsubscript{1.15} & 93.8\textsubscript{1.04} & 89.4\textsubscript{1.20} \\
CLAM-MB\cite{lu2021data} 
& 91.1\textsubscript{0.82} & 94.5\textsubscript{0.78} & 90.7\textsubscript{0.91}
& 83.6\textsubscript{1.42} & 84.8 \textsubscript{0.71} & 81.3\textsubscript{1.70} 
& 89.3\textsubscript{1.23} & 94.2\textsubscript{1.18} & 88.2\textsubscript{1.42} 
& 91.2\textsubscript{0.78} & 92.9\textsubscript{0.66} & 90.2\textsubscript{0.74}  \\
CLAM-SB\cite{lu2021data} 
& 91.9\textsubscript{1.58} & 94.3\textsubscript{1.27} & 91.1\textsubscript{1.54}
& 83.9\textsubscript{1.48} & 85.2\textsubscript{1.64} & 81.5\textsubscript{1.44}
& 87.3\textsubscript{1.23} & 93.1\textsubscript{1.41} & 89.1\textsubscript{1.64} 
& 89.7\textsubscript{1.76} & 93.9\textsubscript{1.67} & 90.2\textsubscript{1.98}  \\
DSMIL \cite{li2021dual} 
& 85.8\textsubscript{0.63} & 91.8\textsubscript{0.72} & 86.2\textsubscript{0.75} 
& 72.2\textsubscript{0.76} & 72.8\textsubscript{0.86} & 72.4\textsubscript{0.72} 
& 85.2\textsubscript{0.85} & 93.6\textsubscript{0.82} & 85.9\textsubscript{0.94} 
& 90.2\textsubscript{0.78} & 94.7\textsubscript{0.66} & 86.2\textsubscript{0.71}  \\
TransMIL \cite{shao2021transmil} 
& 87.8\textsubscript{3.24} & 93.7\textsubscript{3.21} & 88.7\textsubscript{3.61} 
& 75.4\textsubscript{4.02} & 74.6\textsubscript{3.77} & 71.7\textsubscript{3.23} 
& 87.9\textsubscript{3.22} & 94.1\textsubscript{3.12} & 88.2\textsubscript{3.40} 
& 91.1\textsubscript{2.56} & 92.5\textsubscript{2.75} & 89.3\textsubscript{2.98} \\
DTFD \cite{zhang2022dtfd} 
& 89.4\textsubscript{0.73} & 92.3\textsubscript{0.92} & 88.4\textsubscript{0.78} 
& 76.3\textsubscript{0.67} & 77.8\textsubscript{0.88} & 75.4\textsubscript{0.82} 
& 86.8\textsubscript{1.04} & 94.7\textsubscript{0.75} & 86.1\textsubscript{0.91} 
& 91.5\textsubscript{0.79} & 95.3\textsubscript{0.85} & 90.8\textsubscript{0.77}  \\
RRT \cite{tang2024feature} 
& 90.9\textsubscript{1.08} & 94.7\textsubscript{1.44} & 90.2\textsubscript{0.88} 
& 78.9\textsubscript{2.22} & 79.5\textsubscript{1.31} & 78.7\textsubscript{1.54} 
& 89.2\textsubscript{1.98} & 94.4\textsubscript{1.74} & 88.5\textsubscript{1.46} 
& 93.3\textsubscript{1.23} & 95.1\textsubscript{1.78} & 91.2\textsubscript{1.45} \\
WiKG  \cite{li2024dynamic} 
& 91.1\textsubscript{1.26} & 94.6\textsubscript{1.20} & 90.8\textsubscript{1.15} 
& 80.3\textsubscript{1.41} & 80.4\textsubscript{1.38} & 77.8\textsubscript{1.20} 
& 89.7\textsubscript{0.96} & 94.6\textsubscript{0.72} & 89.3\textsubscript{1.23} 
& 93.2\textsubscript{1.11} & 95.9\textsubscript{0.84} & 91.6\textsubscript{1.12}  \\
\textbf{AMD-MIL} & \textbf{92.9\textsubscript{2.73}} & \textbf{96.4\textsubscript{2.89}} & \textbf{92.7\textsubscript{2.83}} & 
\textbf{85.0\textsubscript{1.32}} & \textbf{85.3\textsubscript{0.69}} & \textbf{82.7\textsubscript{1.24}} & 
\textbf{90.5\textsubscript{1.51}} & \textbf{95.2\textsubscript{0.70}}& \textbf{90.5\textsubscript{1.59}}& 
\textbf{94.4\textsubscript{1.13}} & \textbf{97.3\textsubscript{0.74}}& \textbf{92.9\textsubscript{1.01}}\\
\bottomrule
\end{tabular}}
\label{big table}
\end{table*}

\textbf{Mask Agent.} 
In this MIL task, most regions of a WSI do not contribute much to the prediction, so a learnable mask is generated by using the trainable threshold to mask the information.
\begin{align} 
\tau = \sigma(p( W_\tau V_A^T) ),
\end{align}
where $W_\tau \in \mathbb{R}^{1 \times D}$, function $p$ is an  
adjustable aggregate function such as mean-pooling, and $\tau$ is the threshold used in Eq. ~\ref{eq:denoise}. 

Calculate the importance of each feature to optimize the important features in the hidden space. The selection of features will have the risk of information loss. To balance important information selection and the original characteristics of the aggregation, we proposed a new module which can be defined as:

\begin{align} 
V_{MD_{ij}} = V_{A_{ij}} \mathbb{I}_{M_{ij}>\tau}+DN_{ij} \label{eq:denoise},
\end{align}

where $M=W_MV_A$ is the threshold matrix to obtain the importance of each feature, and $DN=W_{DN}V_A$ is the denoise matrix to aggregate information. Here, $W_M$ and $W_{DN}$ are learnable parameters.
\\ %
\textbf{Agent Visualization.}  The foundational agent attention architecture lacks the capability to produce a variable concentration score for sequences. To address this limitation, we outline a methodology that facilitates the visualization of attention scores:

\begin{align} 
Att_i = \sum_{j=1}^n Q_{A_{0,j}} K_{A_{j,i+1}}, \label{eq:vis}
\end{align}
where $Att_i$ is the attention score of the feature $h_i$.
\\ %
\textbf{AMD.} Establishing the aforementioned modules, we introduce a novel framework titled mask denoise mechanism. This framework, as illustrated in Figure~\ref{F2}, encompasses a learning-based agent attention mechanism, representation refinement, and feature aggregation. The algorithm process is shown in Algorithms ~\ref{alg:AOA} and the module can be expressed as:

\begin{equation}
\begin{aligned} 
O &= \sigma(QA^T)V_{MD},
\end{aligned}
\end{equation}
where $md$ represents the mask denoise mechanism, and $V_{MD}$ represents the matrix calculated from the mechanism Eq. ~\ref{eq:denoise}. 

Due to the difference in the threshold selection method, the other two feature threshold selection strategies are considered as follows: 

\begin{itemize}[leftmargin=*]
  \item \textbf{Mean-AMD}. Mean selection: selected the average value in the features as the threshold selected by all features.
  \item \textbf{CNN-AMD}. CNN selection: through the method of group convolution, the characteristics of different groups are reduced, and the average value between the groups is the threshold.
\end{itemize}

\section{Experiments}
\subsection{Datasets and Evaluation Metrics}
In our study, We use four public datasets to assess our approach.
\\ %
\textbf{CAMELYON-16} is a dataset for early-stage breast cancer lymph
 node metastasis detection. The dataset comprises 399 WSIs, which
 are officially split into 270 for training and 129 for testing. We use 6-fold cross-validation to ensure that all data are utilized
 for both training and testing, thereby preventing overfitting to the
 official test set. In addition, we employ the pre-trained weights from
 the CAMELYON-16 dataset to perform inference on the external
 dataset CAMELYON-17 only once. Subsequently, we report both
 the mean and variance of the evaluation metrics.
\\ %
\textbf{TCGA-LUNG} includes 1034 WSIs: 528 from LUAD and 507 from LUSC cases. We split the dataset into 65:10:25 for training, validation, and testing. 4-fold cross-validation is used, and the mean and standard deviation of metrics are reported.
\\ %
\textbf{TCGA-KIDNEY} includes 1075 WSIs: 117 from KICH, 539 from KIRC, and 419 from KIRP cases. We split the dataset into 65:10:25 for training, validation, and testing. We use 4-fold cross-validation and report the mean and standard deviation of metrics.

We report the mean and standard deviation of the macro F1 score, the AUC for one-versus-rest, and the slide-level accuracy (ACC).

\begin{figure*}[htbp]
\centering
\includegraphics[width=1\linewidth]{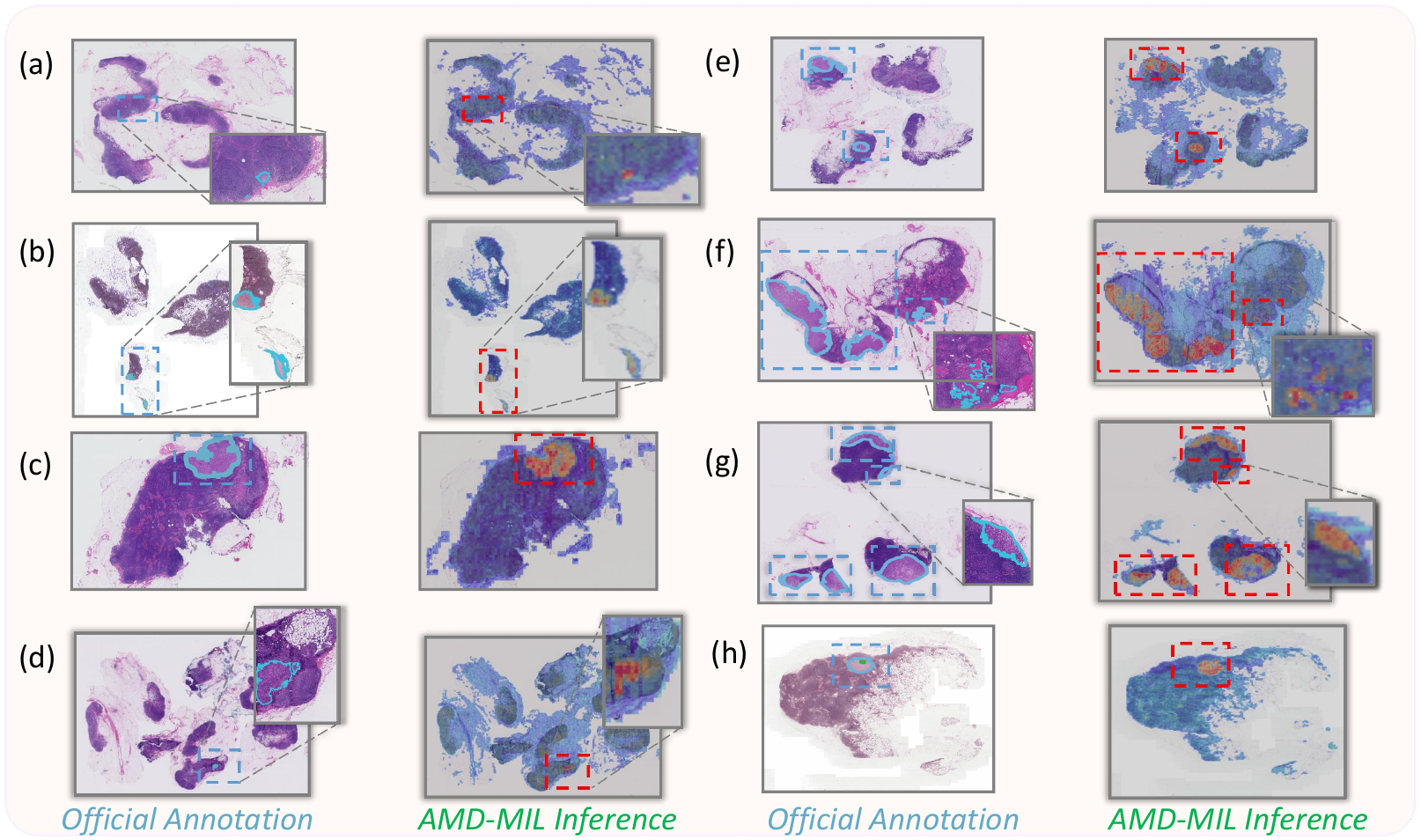}
\caption{Visualization of AMD-MIL Attention Distribution Compared to Official Annotations on CAMELYON dataset.}
\label{F3}
\end{figure*}

\subsection{Implementation Details}
During the pre-processing phase, we generate non-overlapping patches of 256x256 pixels at 20x magnification for the datasets CAMELYON-16, CAMELYON-17, TCGA-KIDNEY, and TCGA-LUNG. This procedure yields an average count of approximately 9024, 7987, 13266, and 10141 patches per bag for the respective datasets. 

Uniform hyperparameters are maintained across all experiments. Each experiment is conducted on a workstation equipped with NVIDIA RTX A100 GPUs, utilizing the ImageNet~\cite{deng2009imagenet} pre-trained ResNet50~\cite{he2016deep} as the feature encoding model. The Adam optimization algorithm is used, incorporating a weight decay of 1e-5. The initial learning rate is set at 2e-4, and cross-entropy loss is employed as the loss function.

\subsection{Comparison with State-of-the-Art Methods}
In this study, we present the experimental results of our newly developed AMD-MIL framework applied to the CAMELYON-16, CAMELYON-17, TCGA-LUNG, and TCGA-KIDNEY datasets. We compare this framework with various methodologies, including MeanMIL, MaxMIL, ABMIL~\cite{ilse2018attention}, CLAM~\cite{lu2021data}, DSMIL~\cite{li2021dual}, TransMIL~\cite{shao2021transmil}, DTFD~\cite{zhang2022dtfd}, RRT~\cite{tang2024feature}, and WiKG~\cite{li2024dynamic}.

As shown in Table~\ref{big table}, the AMD-MIL framework demonstrates superior performance, achieving AUC scores of 96.4$\%$ for CAMELYON-16, 85.3$\%$ for CAMELYON-17, 95.2$\%$ for TCGA-LUNG, and 97.3$\%$ for TCGA-KIDNEY. Notably, these scores consistently exceed those of the previously mentioned comparative methods, highlighting the framework's exceptional ability to dynamically adapt to inputs. This adaptability enables the effective capture of key features, accurately representing the original bag features.

As demonstrated in Table~\ref{thresh table}, we also conduct a comparative analysis to evaluate the impact of different threshold selection methods on the metrics. We find that using a linear layer for aggregation outperforms both average pooling and group convolution.
\subsection{Interpretability Analysis}
We conduct an interpretability analysis of AMD-MIL. In Figure~\ref{F3}, the blue-annotated areas denote the official annotations of cancerous regions in the CAMELYON dataset, whereas the heatmap regions represent the distribution of agent attention scores across all patches constituting the WSIs, calculated according to Eq. ~\ref{eq:vis}. The attention scores indicate the contribution level of instances to the classification outcome, and it is distinctly observable that areas with high attention scores align closely with the annotated cancerous regions. This demonstrates that the AMD-MIL classification relies on cancerous ROIs, mirroring the diagnostic process of pathologists and thereby providing substantial interpretability for clinical applications. AMD-MIL possesses robust localization capabilities for both macro-metastases and micro-metastases. For example, in Figure~\ref{F3} (f), which includes both macro and micro-metastases, AMD-MIL can also concurrently localize to different areas.
\subsection{Ablation Study}

\begin{table*}[htbp]
  \centering
  \tabcolsep=0.32cm
  \caption{ Comparison between TransMIL with AMD-MIL and the effectiveness of the components of AMD-MIL.}
  \begin{tabular}{ccccccccccccc}
    \toprule
    \multirow{2}{*}[-3pt]{Dataset} & \multicolumn{5}{c}{Component} & \multirow{2}{*}[-3pt]{ACC(\%)} & \multirow{2}{*}[-3pt]{AUC(\%)} & \multirow{2}{*}[-3pt]{F1(\%)} \\ 
    \cmidrule(r){2-6}
    & Nystr\"om  & agent  & train & mask & denoise  \\ \hline
    \multirow{3}{*}[-12pt]{CAMELYON-16} 
    & \checkmark &  &  &  & &87.8\textsubscript{3.24} &93.7\textsubscript{3.21} &88.7\textsubscript{3.61} \\ 
    &  & \checkmark  &   & & & 89.3\textsubscript{2.90}& 93.8\textsubscript{3.08}&88.6\textsubscript{3.09} & \ \\ 
    &  & \checkmark  &\checkmark & & & 91.5\textsubscript{3.62}& 95.6\textsubscript{3.06}&91.2\textsubscript{3.67} & \\ 
    &  & \checkmark  &\checkmark &\checkmark & & \textbf{93.0\textsubscript{2.72}} &96.0\textsubscript{3.00} &92.7\textsubscript{2.80} &  \\ 
    &  & \checkmark &\checkmark &\checkmark &\checkmark & 92.9\textsubscript{2.73} & \textbf{96.4\textsubscript{2.89}}& \textbf{92.7\textsubscript{2.83}}&  \\ \hline
    \multirow{3}{*}[-12pt]{LUNG} 
    & \checkmark &  &  &  & &87.9\textsubscript{3.22} &94.1\textsubscript{3.12} &88.2\textsubscript{3.40} \\ 
    &  & \checkmark  &   & & & 88.4\textsubscript{0.86}& 93.5\textsubscript{0.86}&88.4\textsubscript{0.88} & \ \\ 
    &  & \checkmark  &\checkmark & & & 87.5\textsubscript{1.00}& 92.6\textsubscript{3.47}&87.4\textsubscript{1.02} & \\ 
    &  & \checkmark  &\checkmark &\checkmark & & 90.2\textsubscript{1.19} &94.6\textsubscript{0.91} &90.2\textsubscript{1.19} &  \\ 
    &  & \checkmark &\checkmark &\checkmark &\checkmark & \textbf{90.5\textsubscript{1.51}} & \textbf{95.2\textsubscript{0.70}}& \textbf{90.5\textsubscript{1.59}}&  \\ \hline
    \multirow{3}{*}[-12pt]{KIDNEY} 
    & \checkmark &  &  &  & &91.1\textsubscript{2.56} &92.5\textsubscript{2.75} &89.3\textsubscript{2.98 } \\ 
    &  & \checkmark  &   & & & 93.7\textsubscript{0.43}& 97.0\textsubscript{0.57}&91.1\textsubscript{0.94} &  \\ 
    &  & \checkmark  &\checkmark & & & 93.7\textsubscript{1.13}& \textbf{97.7\textsubscript{0.57}}&91.4\textsubscript{0.18} & \\ 
    &  & \checkmark  &\checkmark &\checkmark & & 93.4\textsubscript{1.06} &97.6\textsubscript{0.57} &90.7\textsubscript{0.13} &  \\ 
    &  & \checkmark &\checkmark &\checkmark &\checkmark & \textbf{94.4\textsubscript{1.13}} & 97.3\textsubscript{0.74}& \textbf{92.9\textsubscript{1.01}}&  \\ 
    \bottomrule
    \end{tabular}
    \label{ablation table}
\end{table*}

\begin{figure}[bp]
\centering
\includegraphics[width=1\linewidth]{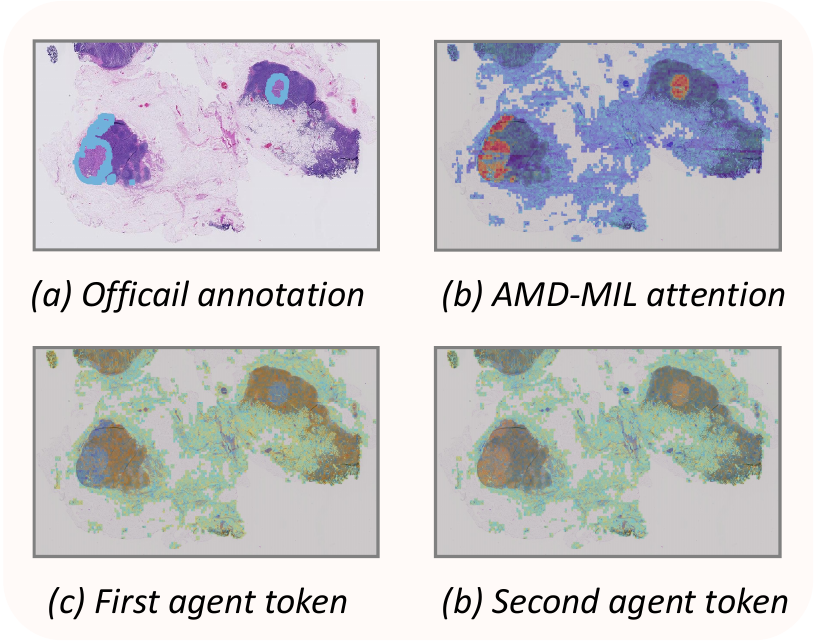}
\caption{Attention distribution of different agent tokens.}
\label{F5}
\end{figure}

\begin{figure}[t]
\centering
\includegraphics[width=0.96\linewidth]{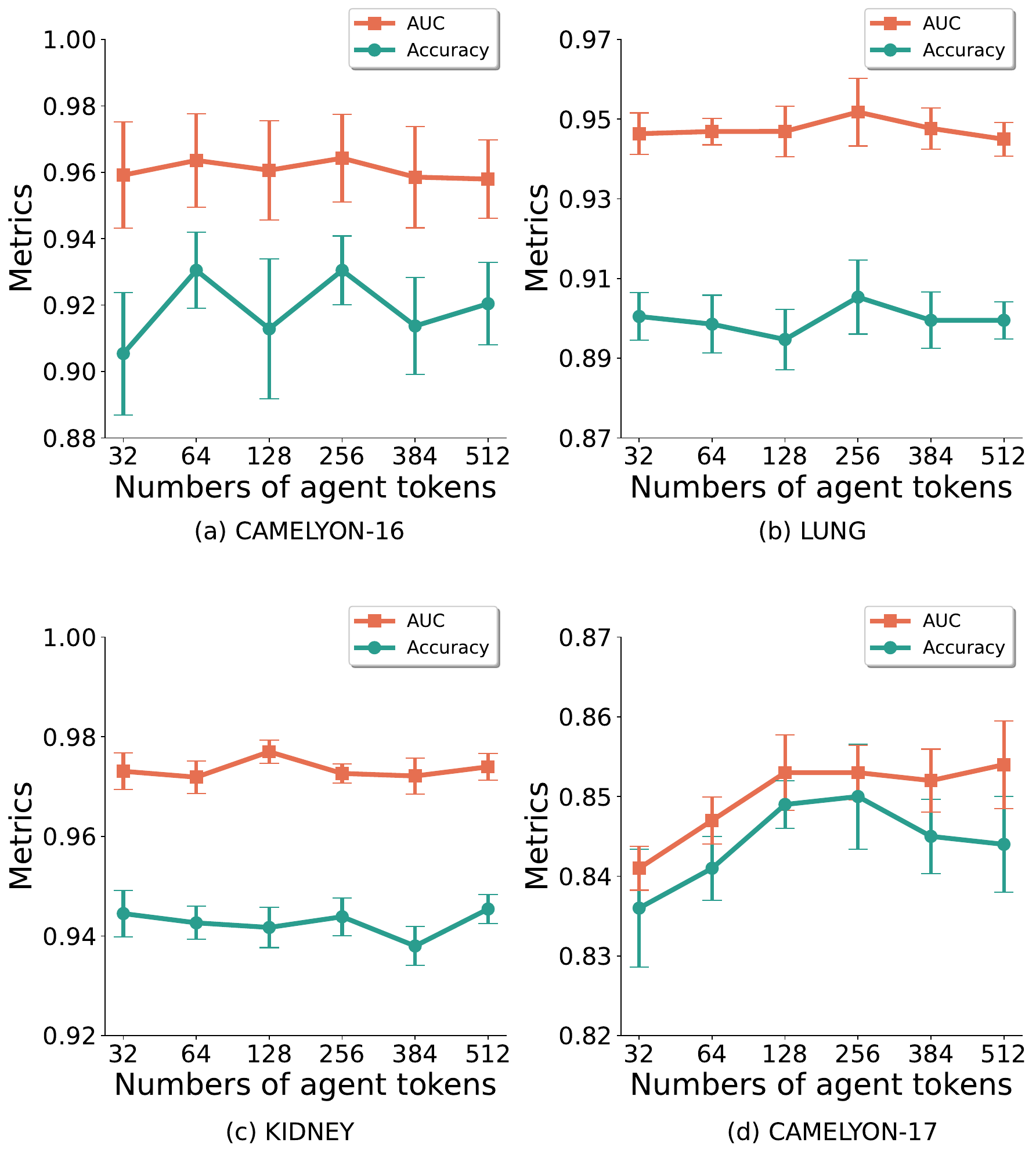}
\caption{Influence of the number of agent tokens}
\label{F4}
\end{figure}

\begin{figure}
\centering
\includegraphics[width=1\linewidth]{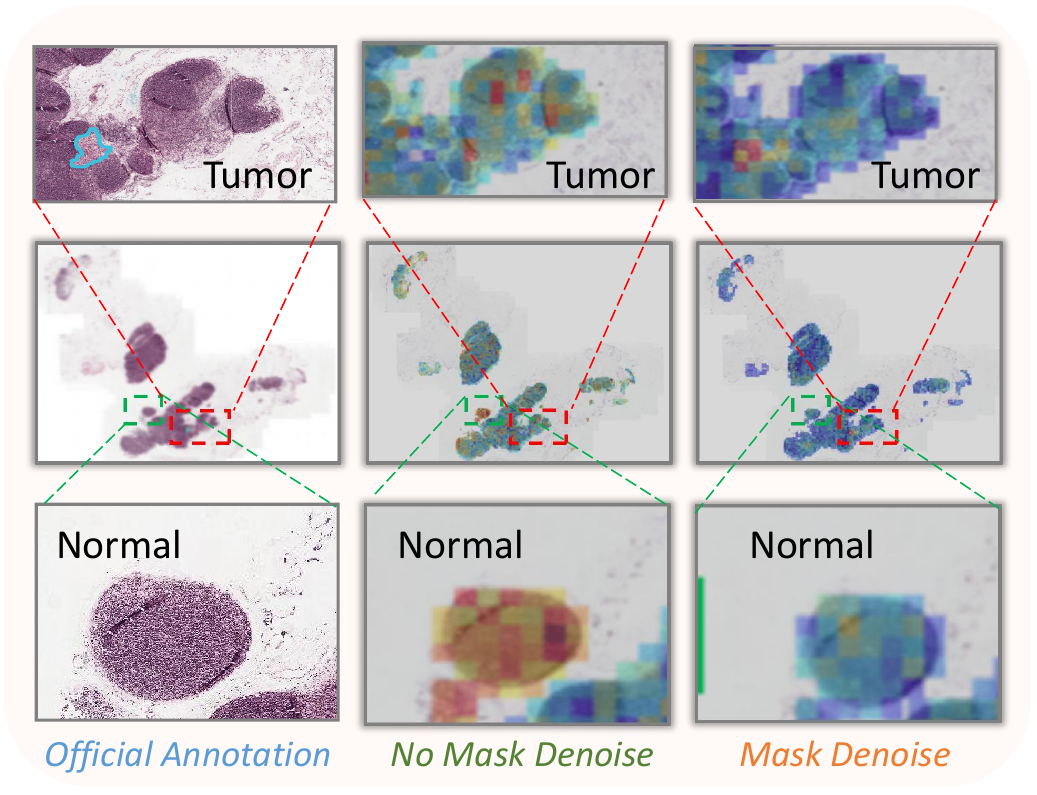}
\caption{Effectiveness of the mask denoise mechanism.}
\label{F6}
\end{figure}

\textbf{Effectiveness of Agent Aggregator.}
The trainable agent aggregator uses agent tokens as intermediate variables for the query and key in the self-attention mechanism. This approach ensures global modeling and approximates linear attention. We compare the trainable agent aggregator with the original pooling agent aggregator and the Nyström attention aggregator from TransMIL. The original pooling agent aggregator reduces parameter count using an agent mechanism and achieves enhanced global modeling through broadcasting. As shown in Table~\ref{ablation table}, it significantly outperforms the Nyström attention aggregator from TransMIL across three datasets.

However, the pooling agent aggregator struggles to adapt dynamically to inputs, and its pooling mechanism may average out important instances. Initializing the agent as a trainable parameter results in improved metrics compared to the pooling agent aggregator. This change allows the model to better adapt to varying inputs and maintain the significance of crucial instances, thereby enhancing overall performance.

We explore the attention distribution patterns among agent tokens as shown in Figure~\ref{F5}. The first agent token focuses on non-cancerous tissues, whereas the second targets cancerous zones. This suggests that different agent tokens have unique focal points. This variance ensures that during broadcasting, diverse queries focus on their respective areas, enhancing the model's ability to differentiate between critical and non-critical regions.

The number of agent tokens is crucial for AMD-MIL performance. Figure~\ref{F4} shows results from experiments on four datasets with agent token counts of 32, 64, 128, 256, 384, and 512. The ACC on the CAMELYON-16 dataset fluctuates with more agent tokens, possibly due to its small size causing instability. On the CAMELYON-17 dataset, the AUC increases with more agent tokens, while the ACC initially increases and then decreases. On the CAMELYON-16 dataset, the AUC and, on the TCGA datasets, both the AUC and ACC maintain stable performance. This consistency aligns with findings from experiments adjusting agent token numbers in shallow attention stacks on natural images~\cite{han2023agent}.
\\ %
\textbf{Effectiveness of Mask Denoise Mechanism.} 
The mask denoise mechanism enhances the allocation of attention scores by selectively masking out less significant representations. Denoising matrices are employed to mitigate noise introduced during the masking process. Table 2 presents a comparison of metrics for the agent aggregator with and without the mask denoise mechanism, demonstrating average performance improvements. Figure 6 illustrates the contrast in the distribution of instance attention scores. Even without the mask denoise mechanism, some WSIs are correctly classified. However, higher attention sometimes targets non-cancerous areas, reducing interpretability. For micro-metastatic cancer, this bias can result in errors, posing significant clinical challenges. With the mask denoise mechanism, attention scores are more focused on cancerous ROIs, thereby reducing attention to non-cancerous regions. This suggests that the mask denoise mechanism enhances interpretability by dynamically correcting attention scores.

\begin{figure}[t]
\centering
\includegraphics[width=1\linewidth]{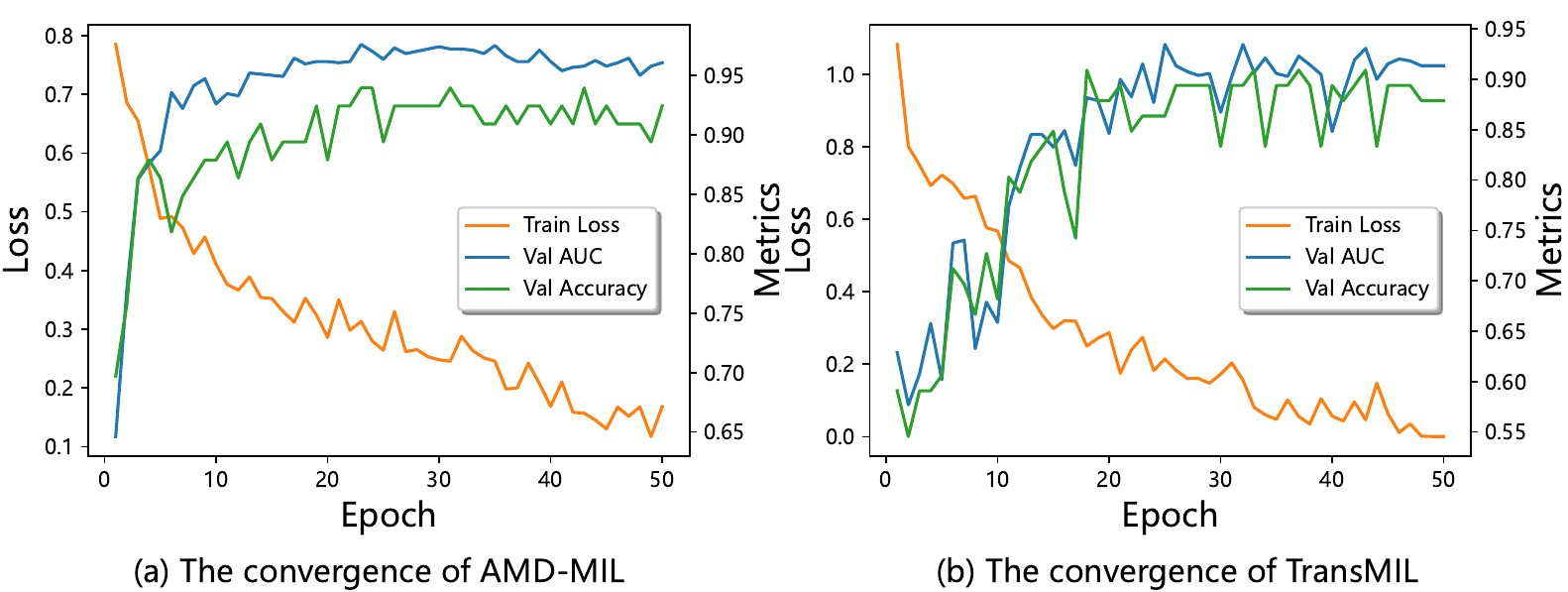}
\caption{Model convergence of AMD-MIL and TransMIL.}
\label{F7}
\end{figure}

As shown in Figure~\ref{F7}, we compare the convergence of AMD-MIL and TransMIL. AMD-MIL shows more stable and better performance.

\begin{table}[htbp]
  \centering
  \tabcolsep=0.39cm 
  \caption{Different thresh select methods on Camelyon-16.}
  \begin{tabular}{cccc}
    \toprule
    Thresh & ACC(\%) & AUC(\%) & F1(\%) \\ 
    \midrule
    Mean  & 91.3\textsubscript{2.12} & 96.0\textsubscript{2.21} & 91.0\textsubscript{3.83} \\
    CNN   & 91.2\textsubscript{3.66} & 95.8\textsubscript{3.39} & 91.0\textsubscript{3.69} \\
    Linear & 92.9\textsubscript{2.73} & 96.4\textsubscript{2.89} & 92.7\textsubscript{2.83} \\
    \bottomrule
  \end{tabular}
  \label{thresh table}
\end{table}

\section*{Conclusion}

In pathological image analysis, attention-based aggregators significantly advance MIL methods. However, traditional attention mechanisms, due to their quadratic complexity, struggle with processing high-resolution images. Additionally, approximate linear self-attention mechanisms have inherent limitations. To address these challenges, we introduce AMD-MIL, a novel approach for dynamic agent aggregation and representation refinement. Our validation on four distinct datasets demonstrates not only AMD-MI's effectiveness but also its ability for instance-level interpretability.

\section*{Acknowledgement}
This study was supported by the Component Project of Shenzhen Pathology Medical Imaging Intelligent Diagnosis Engineering Research Center (XMHT20230115004) and Jilin Fuyuan Guan Food Group Co., Ltd. This work was also supported by the Science and Technology Foundation of Shenzhen City grant number KCXFZ202012211173207022.

\bibliographystyle{ACM-Reference-Format}
\balance
\bibliography{citing}

\end{document}